\pdfoutput=1

\documentclass[11pt]{article}

\usepackage[]{acl}

\usepackage{times}
\usepackage{latexsym}

\usepackage[T1]{fontenc}

\usepackage[utf8]{inputenc}

\usepackage{microtype}

\usepackage{xspace}
\usepackage{amsmath,amsfonts,amssymb}
\usepackage{bm}
\usepackage{multirow}
\usepackage{subfigure}
\usepackage{varwidth}
\usepackage{array}
\usepackage{makecell}

\usepackage{graphicx}
\usepackage{xspace}

\newcommand{\eat}[1]{}
\newcommand{\ie}{\emph{i.e.,}\xspace}
\newcommand{\eg}{\emph{e.g.,}\xspace}

\newcommand{\kw}[1]{{\ensuremath {\mathsf{#1}}}}

\newcommand{\myvec}{\bm}
\newcommand{\mymat}{\textbf}
\newcommand{\mytran}{^\intercal}

\newcommand{\softmax}{\kw{softmax}}
\newcommand{\sparsemax}{\kw{sparsemax}}

\newcommand{\sstfive}{{\sc SST-5}\xspace}
\newcommand{\yelp}{{\sc Yelp-5}\xspace}
\newcommand{\acc}{{\sc Acc}\xspace}

\newcommand{\modelname}{\kw{TraceNet}\xspace}  
\newcommand{\modelminus}{\kw{TraceNet}$^-$\xspace}  
\newcommand{\cnnstatic}{\kw{CNN}-\kw{static}\xspace}  
\newcommand{\cnnrandom}{\kw{CNN}-\kw{rand}\xspace}  
\newcommand{\cnnmulch}{\kw{CNN}-\kw{mulch}\xspace}  
\newcommand{\cnnnons}{\kw{CNN}-\kw{no}\kw{stat}\xspace}  
\newcommand{\lstm}{\kw{LSTM}\xspace}  
\newcommand{\bilstm}{\kw{BiLSTM}\xspace}  
\newcommand{\treelstm}{\kw{GT}-\kw{LSTM}\xspace}  
\newcommand{\bert}{\kw{BERT}\xspace}
\newcommand{\xlnet}{\kw{XLNet}\xspace}  
\newcommand{\roberta}{\kw{RoBERTa}\xspace}  
\newcommand{\glove}{\kw{GloVe}\xspace}  

\title{\modelname: Tracing and Locating the Key Elements in Sentiment Analysis}

\author{Qinghua Zhao, Shuai Ma \\
        SKLSDE Lab, Beihang University, Beijing, China\\
        \{zhaoqh, mashuai\}@buaa.edu.cn}

\begin{document}
\maketitle
\begin{abstract}
We study sentiment analysis task where the outcomes are mainly contributed by a few key elements of the inputs. 
Motivated by the two-streams hypothesis, we explore processing input items and their weights separately by developing a neural architecture, named \modelname, to address this type of task. It not only learns discriminative representations for the target task via its encoders, but also traces key elements at the same time via its locators. 
In \modelname, both encoders and locators are organized in a layer-wise manner, and a smoothness regularization is employed between adjacent encoder-locator combinations. Moreover, a sparsity constraint is enforced on locators for tracing purposes and items are proactively masked according to the item weights output by locators.
A major advantage of \modelname is that the outcomes are easier to understand, since the most responsible parts of inputs are identified. Also, under the guidance of locators, it is more robust to attacks due to its focus on key elements and the proactive masking training strategy. Experimental results show its effectiveness for sentiment classification. Moreover, we provide several case studies to demonstrate its robustness and interpretability. The code and data are released at \url{https://github.com/lshowway/tracenet}.                                                
\end{abstract}

\section{Introduction} \label{sec-intro}
As we all know, in sentiment analysis (SA) task~\cite{chen2019transfer,johnson2015effective,zhang2018deep}, its overall sentiment always depends to a large extent on a few key elements of the inputs.
For example. Given a short movie review ``\textit{deflated ending aside, there's much to recommend the film}'' obtained from the \sstfive dataset (detailel in later Section), the three words \textit{deflated}, \textit{much}, and \textit{recommend} have larger impacts on the overall sentiment polarity of the review.

For this type of task, a lesson from attention mechanism~\cite{bahdanau14neural,vaswani2017attention,velickovic18gat} is worthy of learning, where a weighted sum over all input items is computed.
Despite its effectiveness, this strategy remains simple and could not fully reveal nor exploit the unique input structure, \ie the existence of a few key elements.
To be specific, the input structure is \textbf{implicitly} modeled, it is unclear whether the structure could enhance the model performance in terms of both prediction effectiveness and, better yet, other promising properties such as evaluation and robustness.
Moreover, the importance weights of both attention models are \textbf{dense}, as a result of which the key elements are not directly revealed.

\begin{figure*}[tbh!]
\centering
\includegraphics[width=.95\textwidth]{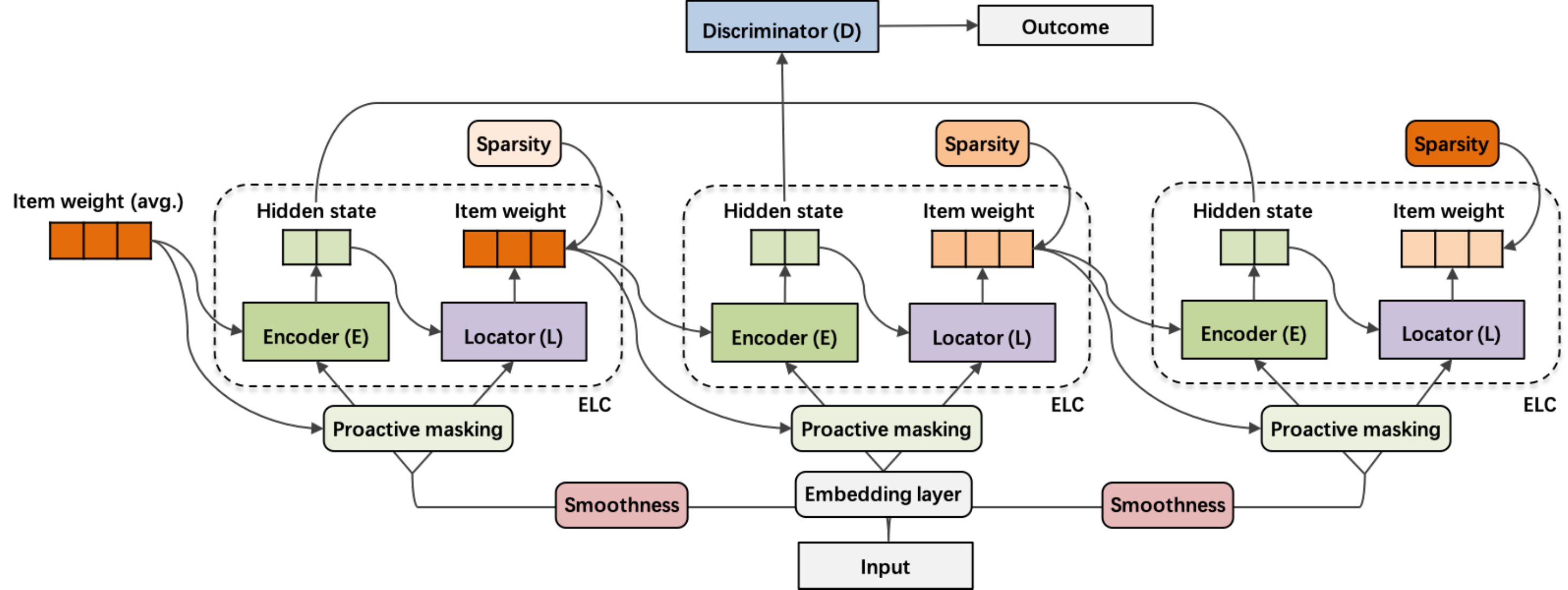}
\caption{General architecture of \modelname (hidden size and the number of input items are 2 and 3).}
\label{fig:framework}
\end{figure*}

To alleviate the above issues and answer the questions, we take one step towards explicitly and separately modeling the input structure. 
Explicitly means that we explicitly associate each input item with a weight and update the weight during the training. Separately means that the input items and item weights are processed separately.
Our work is motivated by the two-streams hypothesis~\cite{goodale1992separate}, which argues that the neural processing of vision and hearing follows two distinct streams. The ventral stream (a.k.a. ``what pathway'') is involved with the object and visual identification and recognition, while the dorsal stream (or, ``where pathway'') is involved with processing the spatial location relative to the viewer and with speech repetition. Such what-and-where decomposition has already shown its usefulness in computer vision~\cite{jacobs1991task, simonyan2014two, 8630333, zhang2021robust} and natural language processing~\cite{zhang2019sentiment} tasks.
We assume that the input structure, i.e., input items and items importance, can be processed by different pathways and then be mutually reinforced. 
To implement this, we explore a neural architecture \modelname, what distinguishes \modelname from previous ones is that it not only learns discriminative representations, but also traces the key input elements at the same time.

Central to \modelname are a set of \underline{E}ncoder-\underline{L}ocator \underline{C}ombination\underline{s} (ELCs) such that encoders and locators are responsible for the ``what and where pathways'' respectively. \modelname adopts a layer-wise architecture to organize ELCs, which enables encoders and locators to collaborate for mutual reinforcement between the two sub-tasks, \ie representation learning and structure revealing. 
More specifically, locators utilize the hidden states of encoders to estimate item weights more accurately, and encoders are in turn guided by the item weights of locators to obtain more discriminative hidden states. 
Also, there is a smoothness regularization between the input item embeddings of adjacent ELCs. This is to prevent the hidden states from changing significantly and ensure the stabilization of learning across layers.
For the purpose of tracing, \modelname further enforces sparsity constraints with increasing strength on locators. As a result, locators are taught to identify a small subset of key elements eventually.
In addition, \modelname employs a proactive masking strategy, \ie proactively masking key elements as indicated by item weights during training. 
The strategy prevents \modelname from simply learning feature co-adaption and assists it to resist attacks on key elements.

We exploit \modelname for SA for evaluation. 
Experimental results on both sentence- and document-level sentiment classification demonstrate the effectiveness of \modelname.
Notably, despite the large-scale training corpus and many engineering efforts for the state-of-the-art pre-trained language models, \modelname built upon \xlnet and \roberta could further increase the classification accuracy over the two.
Then, we provide a case study by considering a total of eight types of attacks, and show that \modelname is more robust to attacks than \xlnet, especially on hard attacks such as changing word orders and dropping information.
Moreover, our qualitative analysis verifies that the revealed item weights make the outcomes of \modelname easier to understand.
Finally, we conduct several experiments to analyse the parameters sensitivity, e.g., masking probability, number of stacked ELCs and hidden state aggregation in each ELCs.

\section{Related Work} \label{sec-related}
\textbf{Word embedding methods.} GloVe~\cite{pennington2014glove} performs on aggregating global word-word co-occurrence statistics from a corpus, it is an unsupervised learning algorithm for obtaining vector representations for words and is publicly available.
 Deep learning models, \eg convolutional neural networks (CNNs) and recurrent neural networks (RNNs), have already demonstrated their superiority for the task~\cite{cho2014learning,choi2018learning,kim2014convolutional}.
Distinct from exploiting the spatial and temporal patterns in texts as done by CNNs and RNNs, \modelname tackles the problem by considering the special input structure such that the outcome is mainly contributed by a few key elements.
Recently, large-scale pre-trained language models~\cite{devlin2019bert,liu2019roberta,Yang2019xlnet} have further led to significant performance gains on a broad range of NLP tasks.
\modelname is capable of integrating any such effort through its embedding layer, and its contribution is to further enhance model performance by tracing key input elements.
While we have also observed a growing trend in aspect-level sentiment analysis~\cite{chen2019transfer,tang2019progressive}, in this work, we only consider the problem at sentence-level and document-level. 

\textbf{Two-stream hypothesis.} 
\cite{zhang2019sentiment} also borrows the notation from the two-stream hypothesis, where the segmentation tagging task is considered as a ``where”-task (i.e., the location of entities), and the sentiment recognition as the ``what''-task. The difference between \modelname and \cite{zhang2019sentiment} is that we separately treat the input items and item weights as ``what'' and ``where'', while the latter considers segmentation tagging and sentiment classification and ``where'' and ``what''. Since there are very different settings and evaluation datasets are adopted, we do not include it as our baseline.

\section{Proposed Model} 
\label{sec-method}

\subsection{General Architecture} \label{subsec-general}

\begin{figure*}[h]
\centering
\includegraphics[width=.95\textwidth]{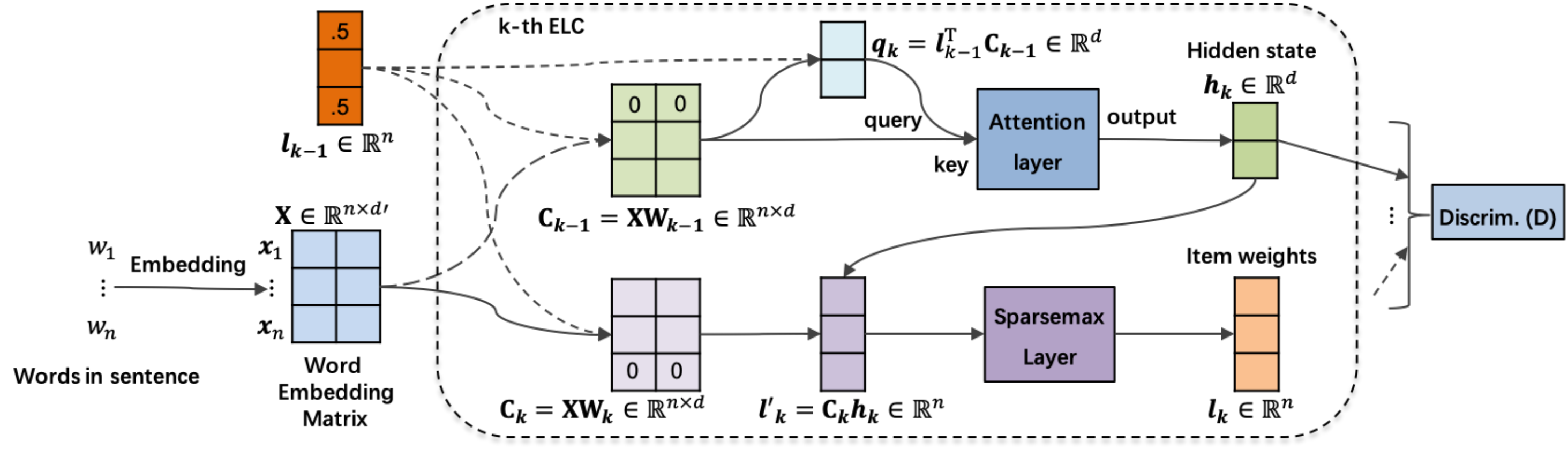}
\caption{Implementation of a single ELC of \modelname ($n=3$, $d'=d=2$ and we omit the bias vectors for computing $\mymat{C}_{k-1}$ and $\mymat{C}_{k}$).}
\label{fig:impl}
\end{figure*}

As mentioned earlier, we consider SA task whose input can be represented as a set of items, and the corresponding outcome is mainly contributed by a few key items.
The proposed model is illustrated in Fig.~\ref{fig:framework}.
\modelname first transforms the item-based input into continuous vector representation in its \textbf{embedding layer}. 
The core of \modelname is a set of \textbf{encoder-locator combinations} (ELCs) organized layer-by-layer, as shown in the vertical-middle part of Fig.~\ref{fig:framework}. Each ELC behaves as a basic functional unit of \modelname, which jointly learns task-specific representation and reveals input structure.
There is a \textbf{smoothness regularization} between the input item embeddings of adjacent ELCs. This is to prevent the hidden states from changing significantly and ensure the stabilization of learning across layers. 
\modelname further places a \textbf{sparsity constraint} on the vector to derive sparse item weights. More specifically, it increases the strength of sparsity constraints on locators layer-by-layer, as shown by the varying colors of the sparsity components in Fig.~\ref{fig:framework}. Since it is generally more challenging to identify key elements at the very beginning, the weaker sparsity constraint allows locators to select more key items for better error tolerance. 
Then the \textbf{proactive masking} strategy masks some input items (\ie setting the corresponding embeddings to zero) during training to boost model performance. As we describe the masking process as ``proactive'', it differs from traditional random masking like in BERT~\cite{devlin2019bert} in the way that the probability of each item to be masked is given by its item weight.
At the top of \modelname is a \textbf{discriminator} ${D}$ built to derive the corresponding outcome of every given input with respect to the task.

\subsection{Input \& Embedding Layer}
For sentiment analysis, the input can be unified as a sequence of words $S = [w_1, w_2, \dots, w_n]$.
The embedding layer could be any pre-trained language models among which BERT~\cite{devlin2019bert}, XLNet~\cite{Yang2019xlnet}, and RoBERTa~\cite{liu2019roberta} are the most effective and popular.
As such, each word $w_i\in S$ is transformed into a continuous vector representation $\myvec{x}_i \in \mathbb{R}^{d'}$, $d'$ represent the dimension of embeddings.
By stacking these word vectors, we also have the corresponding word embedding matrix $\mymat{X}\in\mathbb{R}^{n\times d'}$.

\subsection{ELC \& Sparsity Constraint}
For the $k$-th ELC ($k < 1$), given the masked $\mymat{C}_{k-1}$ and $\myvec{l}_{k-1}$, the encoder essentially derives the hidden state $\myvec{h}_{k}\in\mathbb{R}^{d}$ by summing over rows/words in $\mymat{C}_{k-1}$ such that those more important are given higher weights. $d$ is the dimension of vector representations.
To achieve this, it first computes a query vector $\myvec{q}_{k} = \myvec{l}_{k-1}\mytran \mymat{C}_{k-1}$, which encodes key items in the current ELC based on the (sparse) item weights in $\myvec{l}_{k-1}$.
Thus, the query vector $\myvec{q}_{k}$ could determine which words the encoder should pay more attention to. The hidden state $\myvec{h}_{k}$ is then outputted by an attention layer, given $\myvec{q}_{k}$ as query and rows in $\mymat{C}_{k-1}$ as keys/values. Formally, the unnormalized attention weights are given by:
\begin{small}
\begin{equation} \label{eq:attention-weight}
a(\myvec{q}_{k}, \myvec{c}_i^{k-1}) = \myvec{v}_k\mytran \tanh(\mymat{W}_k^{att,q} \myvec{q}_{k} + \mymat{W}_k^{att,c} \myvec{c}_i^{k-1} + \myvec{b}_k^{att}),
\end{equation}
\end{small}
where $\myvec{c}_i^{k-1}$ is the $i$-th row of $\mymat{C}_{k-1}$. Again, $\mymat{W}_k^{att,q} \in \mathbb{R}^{d\times d}$, $\mymat{W}_k^{att,c} \in \mathbb{R}^{d\times d}$, $\myvec{v}_k \in \mathbb{R}^{d}$ and $\myvec{b}_k^{att} \in \mathbb{R}^d$ are learnable parameters in the $k$-th ELC. Finally, hidden state $\myvec{h}_k$ is computed by:
\begin{equation} \label{eq:hidden-state}
\myvec{h}_k = \sum_i \frac{\exp(a(\myvec{q}_{k}, \myvec{c}_i^{k-1}))}{\sum_j \exp(a(\myvec{q}_{k}, \myvec{c}_j^{k-1}))} \myvec{c}_i^{k-1}.
\end{equation}

As for the locator to update item weights, it first obtains the \textit{dense} item weight vector $\myvec{l}'_k = \mymat{C}_{k}  \myvec{h}_k \in \mathbb{R}^n$ based on the masked $\mymat{C}_{k}$ and new hidden state $\myvec{h}_k$. We adopt the \sparsemax\ activation~\cite{martins2016sparsemax} to provide sparsity for $\myvec{l}'_k$. More specifically, $\sparsemax(\myvec{l}'_k)$ returns the euclidean projection of $\myvec{l}'_k$ on the probability simplex of the $n$-dimensional space. By this definition, the sparsity strength of \sparsemax\ is not controllable. On the other hand, the activation of \sparsemax\  depends ultimately on the absolute difference between the values in $\myvec{l}'_k$. Intuitively, the lower the absolute difference is, the less sparse the activation is. We thus turn to linearly scaling $\myvec{l}'_k$ before computing \sparsemax:
\begin{equation}  \label{eq:sparsemax-rev}
\myvec{l}_k = \sparsemax(\sigma(-\sum_{j=k}^{L-1} w_j^2 + w_L^2) \cdot \myvec{l}'_k),
\end{equation} 
where $L$ is the number of layers in \modelname, $\sigma(x) = 1/(1+\exp(-x)) \in (0,1)$ is the sigmoid function, and $w_j\in\mathbb{R}$ $(1\le j \le L)$ are learnable parameters.
As can be easily verified, the linearly scaling weights increase with the increment of $k$, resulting in the increasing strength of sparsity.

\subsection{Smoothness Regularization}
After performing the proper transformation, the word embedding matrix $\mymat{X}$ is fed into encoders and locators repetitively for further learning. 
To obtain layer-wise smoothness, we adopt the adjacent weight tying approach~\cite{Fung2018mem2seq,sukhbaatar2015end}. 
Recall that each ELC requires two distinct transformed word embedding matrices that are used by the inside encoder and locator, respectively.
The main idea of adjacent weight tying is to let every two adjacent ELCs share one transformed word embedding matrix. 
Formally, the $k$-th ELC ($k > 1$) only requires a newly-transformed matrix $\mymat{C}_k=\mymat{X}\mymat{W}_k + \myvec{b}_k \in \mathbb{R}^{n \times d}$ (the solid arrow from $\mymat{X}$ to  $\mymat{C}_k$ in Fig.~\ref{fig:impl}) and re-uses $\mymat{C}_{k-1}=\mymat{X}\mymat{W}_{k-1}+\myvec{b}_{k-1} \in \mathbb{R}^{n \times d}$ from the previous ELC (the dashed arrow from $\mymat{X}$ to  $\mymat{C}_{k-1}$ in Fig.~\ref{fig:impl}). Here $\mymat{W}_k \in \mathbb{R}^{d' \times d}$ and $\myvec{b}_{k} \in \mathbb{R}^{d}$ are learnable parameters in the $k$-th ELC.
As for the first ELC, two transformed word embedding matrices are still required.

\subsection{Proactive Masking}
Before the core computation in the $k$-th ELC, $\mymat{C}_{k-1}$ and $\mymat{C}_{k}$ are further pre-processed by masking with a fixed probability. Take $\mymat{C}_{k-1}$ as an example. With a pre-defined probability $P_{msk}$, $\mymat{C}_{k-1}$ will be masked. 
We perform independent Bernoulli experiments for each row of $\mymat{C}_{k-1}$ and the success rate of each experiment is equal to the corresponding item weight in $\myvec{l}_{k-1} \in \mathbb{R}^{n}$ ($\myvec{l}_{k-1}$ is an input to the $k$-th ELC).
Afterward, all rows that pass the Bernoulli experiments will be replaced with zero. 
Note that this step is only turned on during training.
Figure~\ref{fig:impl} also illustrates an example of proactive masking. Assume vector $\myvec{l}_{k-1} = [0.5, 0, 0.5]\mytran$ and $P_{msk}=1$. Thus, both $\mymat{C}_{k-1}$ and $\mymat{C}_{k}$ are to be masked. For $\mymat{C}_{k-1}$, it turns out only the first row passes the experiment, resulting in the first row being replaced with zero. Similarly, the last row of $\mymat{C}_{k}$ passes the experiment and we show the masked $\mymat{C}_{k}$ in Fig.~\ref{fig:impl}.

\subsection{Discriminator}
We simply adopt a single layer feedforward neural network given the mean of all hidden states to build the discriminator:
\begin{small}
\begin{equation} \label{eq:discriminator}
D([w_1,w_2,\dots,w_n]) = \softmax((\frac{1}{k}  \sum_k \myvec{h}_k) \mymat{W}^{dis}  + \myvec{b}^{dis}).
\end{equation}
\end{small}
Here, $D([w_1,w_2,\dots,w_n])$ is the predictive sentiment class of the input. Assuming the number of classes being $C$, we have learnable parameters $\mymat{W}^{dis} \in \mathbb{R}^{d\times C}$ and $\myvec{b}^{dis} \in \mathbb{R}^{C}$.

\section{Experiments} \label{sec-exp}

\begin{table*}[ht!]
\centering
\label{table:overall}
\begin{small}
\begin{tabular}{c | c c c c | c c c c} \hline
\multirow{4}{*}{\sstfive} 
& \cnnrandom & 39.46 & \lstm & 45.04 & \bert & 51.99 & \modelminus-X & 54.86 \\
& \cnnstatic & 44.32 & \bilstm & 45.18 & \xlnet & 55.20 & \modelname-X & 55.55 \\
& \cnnnons & 44.62 & \treelstm & 40.70 & \roberta & 56.49 & \modelminus-R & 56.59\\
& \cnnmulch & 43.54 & \modelname-G & \bf{46.33} & & & \modelname-R & \bf{57.34} \\ \hline
\multirow{4}{*}{\yelp} 
& \cnnrandom & 56.38 & \lstm & 57.14 & \bert & 63.42 & \modelminus-X & 66.89 \\
& \cnnstatic & 56.30 & \bilstm & 55.32 & \xlnet & 66.75 & \modelname-X & 67.23 \\
& \cnnnons & 57.24 & \treelstm & 53.38 & \roberta & 67.66 & \modelminus-R &  66.92\\
& \cnnmulch & 57.14 & \modelname-G & \bf{58.68} & & & \modelname-R & \bf{67.70} \\ \hline
\end{tabular} 
\caption{Overall accuracy (\%) of sentiment classification.}
\end{small}
\end{table*}

\subsection{Experimental Setting}  \label{subsec-setting}

\textbf{Datasets}. We chose two datasets (\sstfive and \yelp) to evaluate our \modelname. 
\begin{itemize}
    \item \sstfive (Stanford Sentiment Treebank)~\cite{socher2013sst5} is a sentence-level sentiment classification with five sentiment classes (\ie very negative, negative, neutral, positive, very positive). We adopted the provided data split, resulting in 8,544, 1,101, and 2,210 sentences in the training, validation, and test sets, respectively. The average length of sentences is 18 words.
    
    \item \yelp is a document-level review corpus released in the Yelp Dataset Challenge 2015. It has five sentiment classes and the full dataset contains approximately 700,000 documents with an average length of 155 tokens. Due to GPU resource limitation, we only tested on a random 5\% sample of the data, resulting in 32,500, 2,500, and 2,500 documents for training, validation, and test, respectively.
\end{itemize}

\textbf{Metric}. We adopted the classification accuracy (\acc) to evaluate performance, which is the fraction of accurately classified test instances over all test instances.

\textbf{Baselines}. We compared \modelname with three types of baselines and one simplified variant.
\begin{itemize}
    \item \cnnrandom, \cnnstatic, \cnnnons, and \cnnmulch are originally proposed in~\cite{kim2014convolutional}. They only differ in word vectors. 
    
    \item \lstm, \bilstm, and \treelstm are RNN-based baselines. We followed the implementation in~\cite{cho2014learning} for Long Short-Term Memory (\lstm) and bidirectional LSTM (\bilstm). Gumble Tree LSTM~\cite{choi2018learning} (\treelstm) is a tree-structured LSTM which further composes task-specific tree structures.

    \item \bert~\cite{devlin2019bert}, \xlnet~\cite{Yang2019xlnet}, and \roberta~\cite{liu2019roberta} are the state-of-the-art pre-trained language models.  \modelname-G, \modelname-X, \modelname-R represent that the output of \glove, \xlnet and \roberta are treated as the input of \modelname, respectively.
\end{itemize}

\textbf{Implementation details}. We used the official implementation of all baselines provided by authors. Pre-trained word vectors for CNN and RNN baselines were obtained from \glove~\cite{pennington2014glove}. We started with the hyper-parameters recommended in the original papers and finetuned them on the validation set. 
Since \bert, \xlnet, and \roberta were sensitive to batch size, learning rate, and maximum length of words on the small \sstfive data, we performed a grid search over $\{16, 32, 64\}$, $\{2\mbox{e}{-5}, 3\mbox{e}{-5}, 5\mbox{e}{-5}\}$, and $\{64, 128, 256\}$ for the three parameters, respectively.
Please refer to the supplementary material for the concrete parameters. Code will be publicly available when the paper is accepted.

\subsection{Main Results}  \label{subsec:results}
In the first set of tests, we evaluate the overall performance of all approaches for sentiment classification. All tests were repeated five times. The average results are reported in Table~1, where the letters after \modelname and \modelminus indicate the different embedding methods, \ie \glove (G), \xlnet (X), and \roberta (R).

\begin{table*}[ht]
\centering
\label{table:robustness}
\begin{small}
\begin{tabular}{c | c c c | c c} \hline
\bf{Attack} & (a) \xlnet & (b) \modelminus-X & (c) \modelname-X & (c)-(a) & (c)-(b) \\ \hline
None & 55.20 & 54.86 & \bf{55.55} & 0.35 & 0.69 \\ \hline
Replacement (cosine) & 52.01 & 51.83 & \bf{52.82}* & \bf{0.81} & 0.99 \\
Replacement (SWN) & 51.11 & 51.46 & \bf{52.34}** & \bf{1.23} & 0.88 \\
Insertion & 47.69 & \bf{48.30} & 48.13 & \bf{0.44} & -0.17 \\
Shuffle & 41.69 & 43.61 & \bf{43.95}** & \bf{2.25} & 0.33 \\
Deletion & 41.89 & 43.19 & \bf{43.73}** & \bf{1.85} & 0.54 \\
Reversing  & 41.67 & 42.99 & \bf{43.39} & \bf{1.72} & 0.40 \\
Replacement (random) & 37.94 & \bf{39.28} & 39.06* & \bf{1.12} & -0.22 \\
Concatenation & 36.56 & 35.93 & \bf{38.96} & \bf{2.40} & 3.03  \\ \hline
\end{tabular} \\
*/**: significantly outperform \xlnet at the 0.05/0.01 level, t-test
\caption{Accuracy (\%) of sentiment classification  under attacks on \sstfive.}
\end{small}
\end{table*}

We first compare \modelname-G with other CNN and LSTM baselines. Except for \cnnrandom, these approaches all exploit \glove for initializing word embeddings and, therefore, can ensure a fair comparison. According to our tests, CNN and LSTM are generally comparable in terms of sentiment classification. By explicitly revealing the input structure, \modelname-G obtains more promising results, which outperforms all approaches on the sentence-level \sstfive data. On the document-level \yelp dataset, we find that LSTMs are better than CNNs and \modelname-G is the best among its counterparts.

The recent large-scale pre-trained language models significantly increase \acc compared with the aforementioned approaches. We also observe a consistent trend in their performance, such that \roberta is the best, followed by \xlnet and \bert. Built upon these efforts, \modelname is able to further enhance the performance. Notably, it refines the results of \xlnet on both datasets. Finally, by comparing \modelname with \modelminus, we find that the proactive masking strategy consistently has a positive impact. 
All the above results verify the effectiveness of \modelname.

\subsection{Analysis Under Attacks}

In the second set of tests, we evaluate the robustness of \modelname under attacks. Here we only experiment on \sstfive  as the sentiment polarities of sentences are easier to be influenced given its shorter average length. We also only consider \xlnet as the embedding method for \modelname since \roberta (named from \underline{R}obustly \underline{o}ptimized \underline{BERT} \underline{a}pproach) has been augmented with a lot of robust designs including training the model longer, with bigger batches over more data, training on longer sequences, etc.\footnote{As such, we admit that \modelname does not exhibit obviously better robustness compared with \roberta.}

We consider eight types of attacks. More specifically,
\textbf{Reversing} and \textbf{Concatenation} are deterministic attacks such that the former reverses the word orders and the latter concatenates all words in a sentence into one (it will be sliced by \xlnet later). The rest are stochastic attacks. The manipulation of \textbf{Shuffle} is clear by its name.
For \textbf{Insertion}, \textbf{Deletion}, and \textbf{Replacement (random)}, we correspondingly modify one-third of words in a sentence and the new words (if needed) are uniformly sampled following the negative sampling method in word2vec~\cite{mikolov2013word2vec}.
Finally, for (a) \textbf{Replacement (cosine)} and (b) \textbf{Replacement (SWN)}, we replace one-third of words in a sentence with (a) their closest terms evaluated by cosine similarity between \glove vectors and (b) alternative terms within the same sentiment groups in SentiWordNet~\cite{baccianella2010swn}.
We trained models on the original training data and computed \acc on the attacked test data.
The results are reported in Table~2 where the numbers for stochastic attacks are the average results of ten independent runs on different attacked test sets.

\eat{
\begin{itemize}  
\item Insertion: inserting another one-third of words into each sentence such that inserted words are uniformly sampled following the negative sampling of word2vec~\cite{mikolov2013word2vec}.
\item Deletion: deleting one-third of words from each sentence
\item Replacement: replacing one-third of the words in each sentence with 
\item Replacement (cosine): replacing one-third of the words in each sentence with the 
\item Replacement (SWN):
\item Shuffle: Shuffling all words in a sentence.
\item Reversing: reversing the word order.
\item Concatenation: concatenating all words in a sentence into one and leaving word slicing to \xlnet.
\end{itemize}
}

The results are arranged in the ascending order of the strength of attacks, as evaluated by the \acc of \modelname.  \textbf{Replacement (cosine)} and \textbf{Replacement (SWN)} are weaker than the other attacks since the semantics or sentiment polarities of terms are not substantially changed.
The following is \textbf{Insertion} which only introduces noises. Changing word orders (\textbf{Shuffle} and \textbf{Reversing}) and dropping information (\textbf{Deletion}) almost tie in terms of attack strength. Finally, the hardest attacks are \textbf{Replacement (random)} and \textbf{Concatenation} which both remove original information and introduce noises.  
Note that the above conclusions should be taken under our attack setting.

Under all attacks, \modelname is consistently  better than \xlnet, further verifying the effectiveness of explicitly revealing the input structure. More importantly, the absolute improvement of \modelname over \xlnet is higher than on original data (\ie 0.35\%), which indicates that \modelname is generally more robust than \xlnet under attacks. Since the \acc decreases under attacks, the relative improvement is indeed more prominent.
Notably, \modelname is good at dealing with harder attacks such as changing word orders and dropping information.

Finally, comparing \modelname with \modelminus, we can conclude that proactive masking boosts model performance in general under attacks. It is especially effective for \textbf{Concatenation} which will drop much information after re-slicing by \xlnet. However, proactive masking could also lead to negative impacts under \textbf{Insertion} and \textbf{Replacement (random)} since it is not optimized for dealing with inserted noises.

\begin{figure*}[ht]
\centering
\subfigure{\label{exp:itemweight-1}
\includegraphics[trim=80 50 80 40,clip,scale=.35]{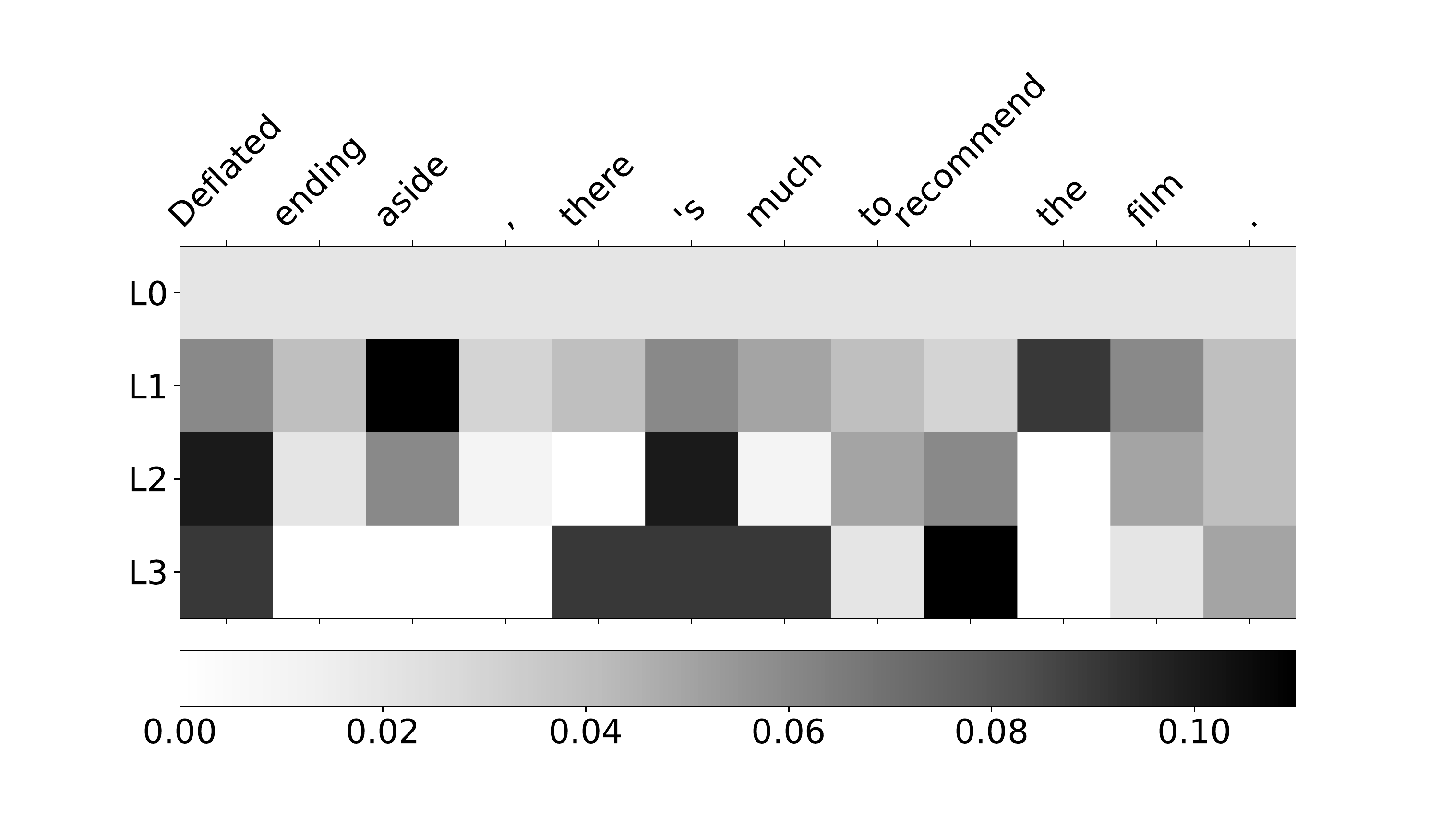}}
\subfigure{\label{exp:itemweight-2}
\hspace{3ex}
\includegraphics[trim=40 50 30 50,clip,scale=.34]{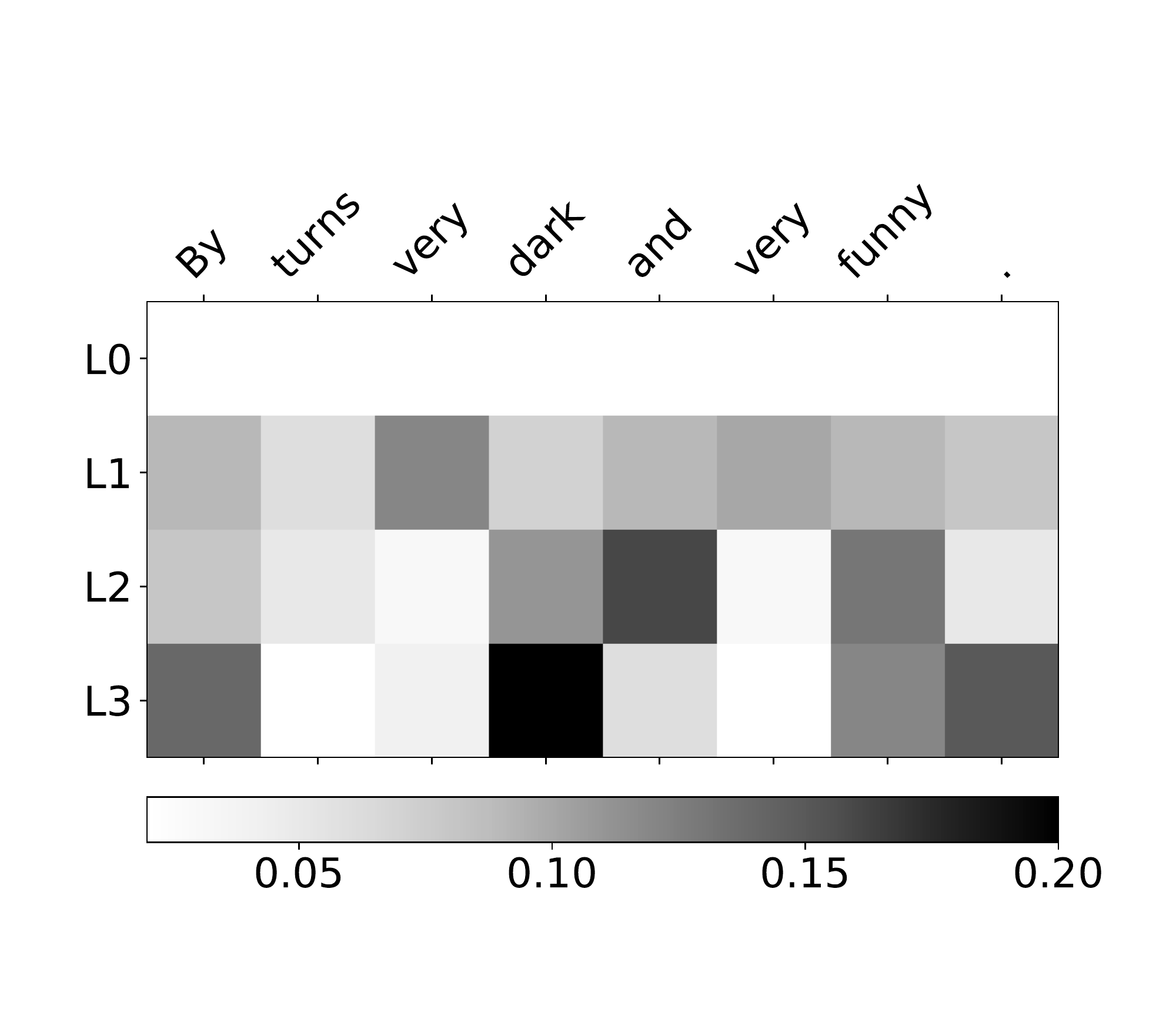}}
\caption{Illustration of item weights identified by \modelname}
\label{exp:itemweight}
\end{figure*}

\begin{figure*}[htb!]
\centering
\subfigure[\sstfive]{\label{exp:pmask-sst}
\includegraphics[width=0.85\columnwidth]{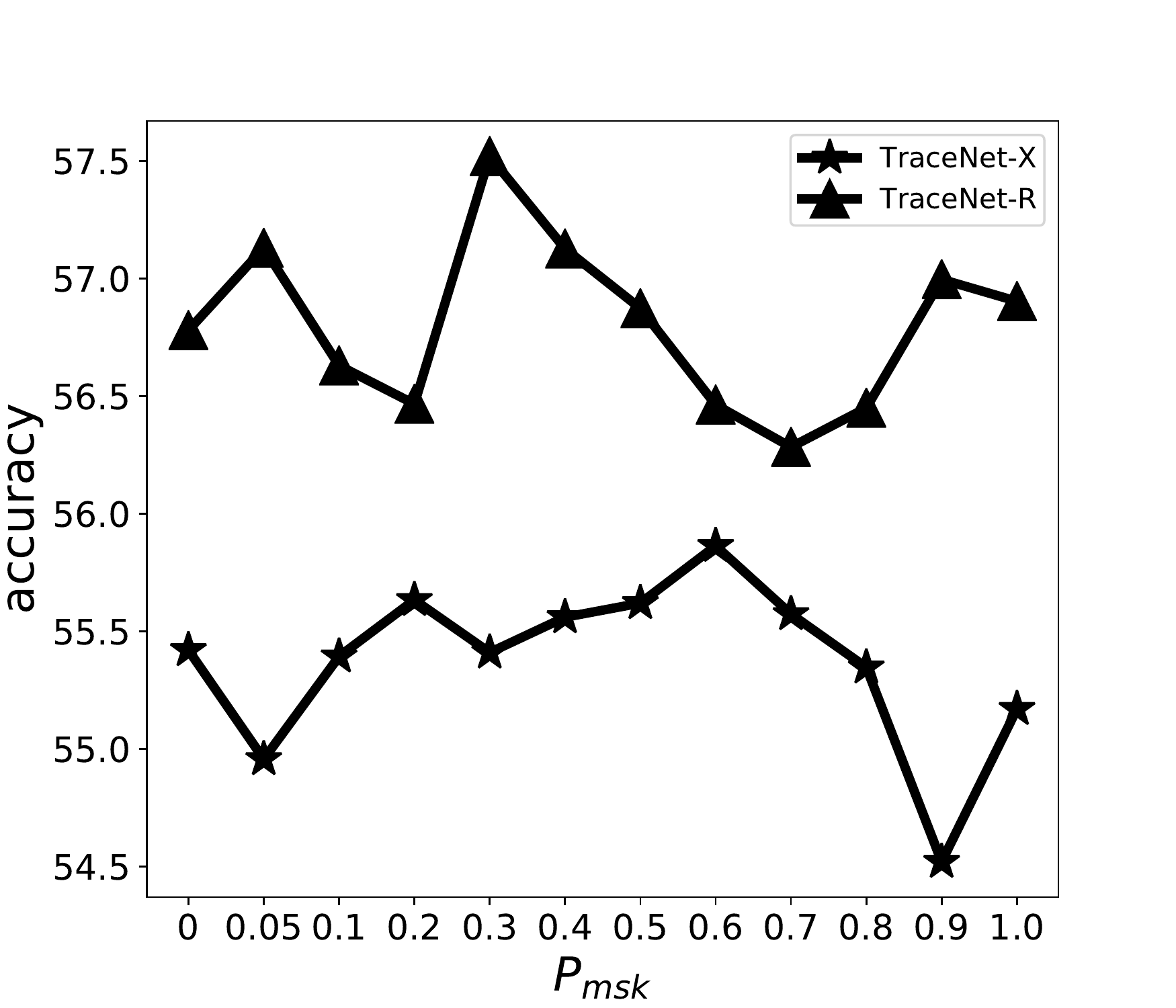}}
\subfigure[\yelp]{\label{exp:pmask-yelp}
\includegraphics[width=0.85\columnwidth]{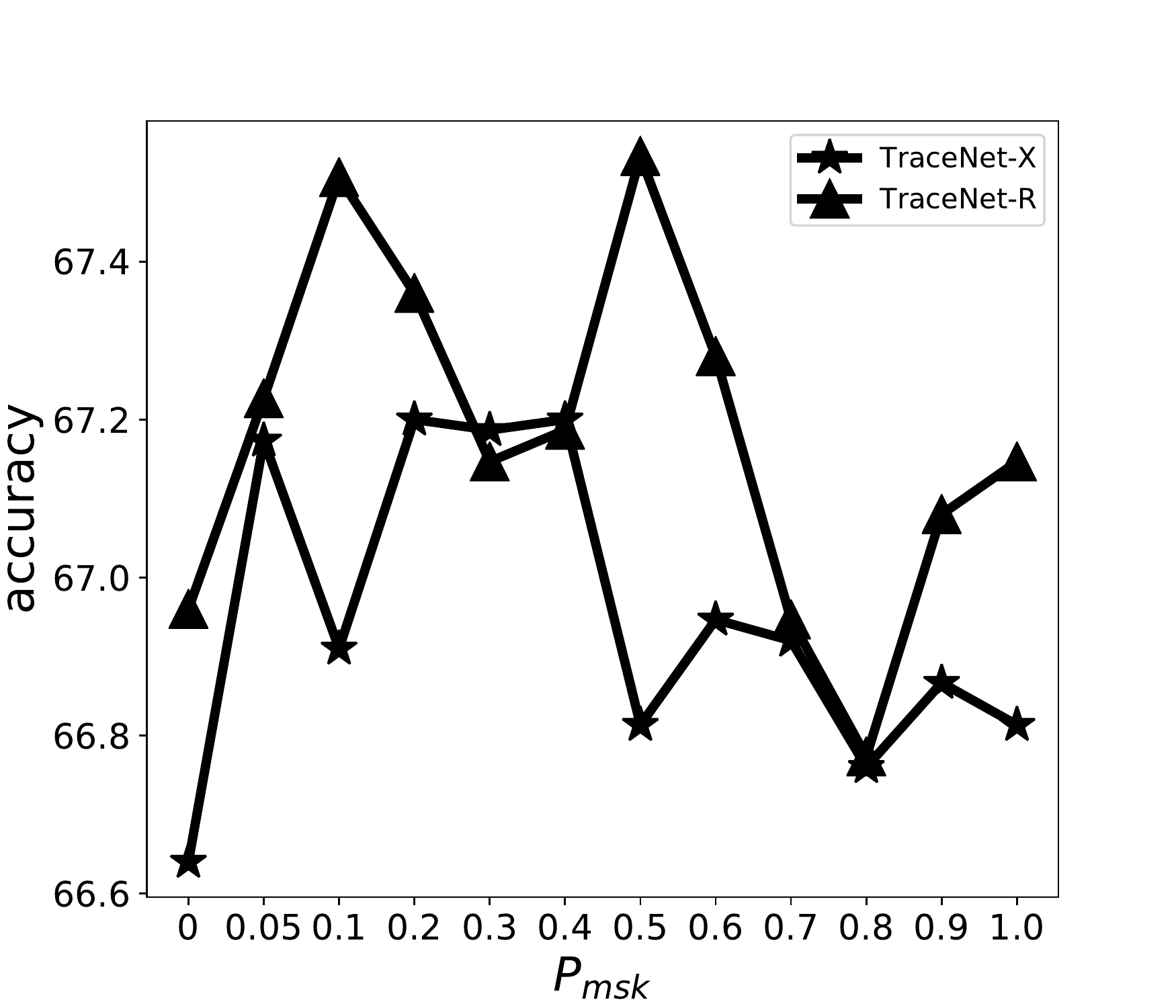}}
\caption{Impacts of masking probability $P_{msk}$.}
\label{exp:sens-pmask}
\end{figure*}

\begin{figure*}[h]
\centering

\subfigure[\sstfive]{\label{exp:num_layer-sst}
\includegraphics[width=0.85\columnwidth]{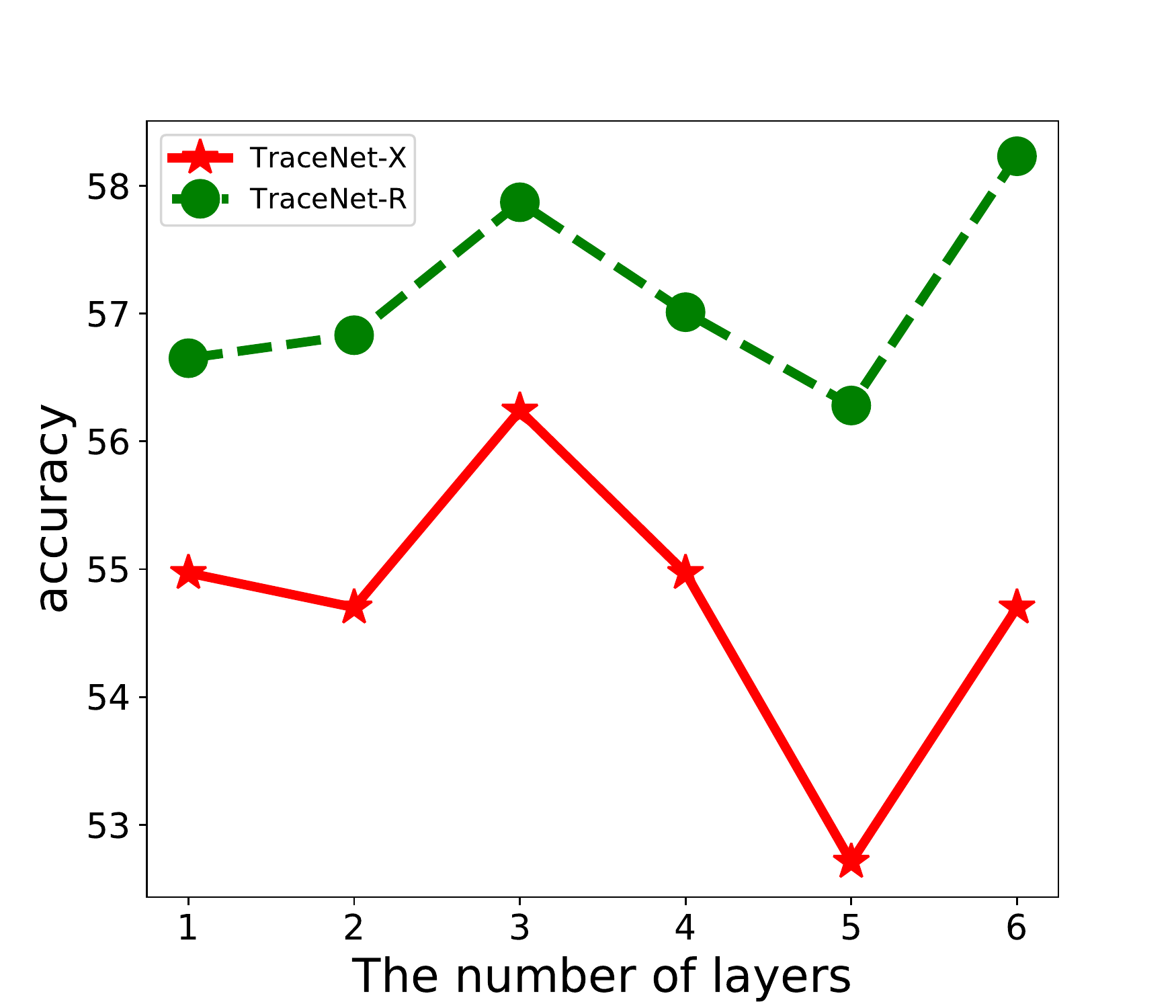}}
\subfigure[\yelp]{\label{exp:num_layer-yelp}
\includegraphics[width=0.85\columnwidth]{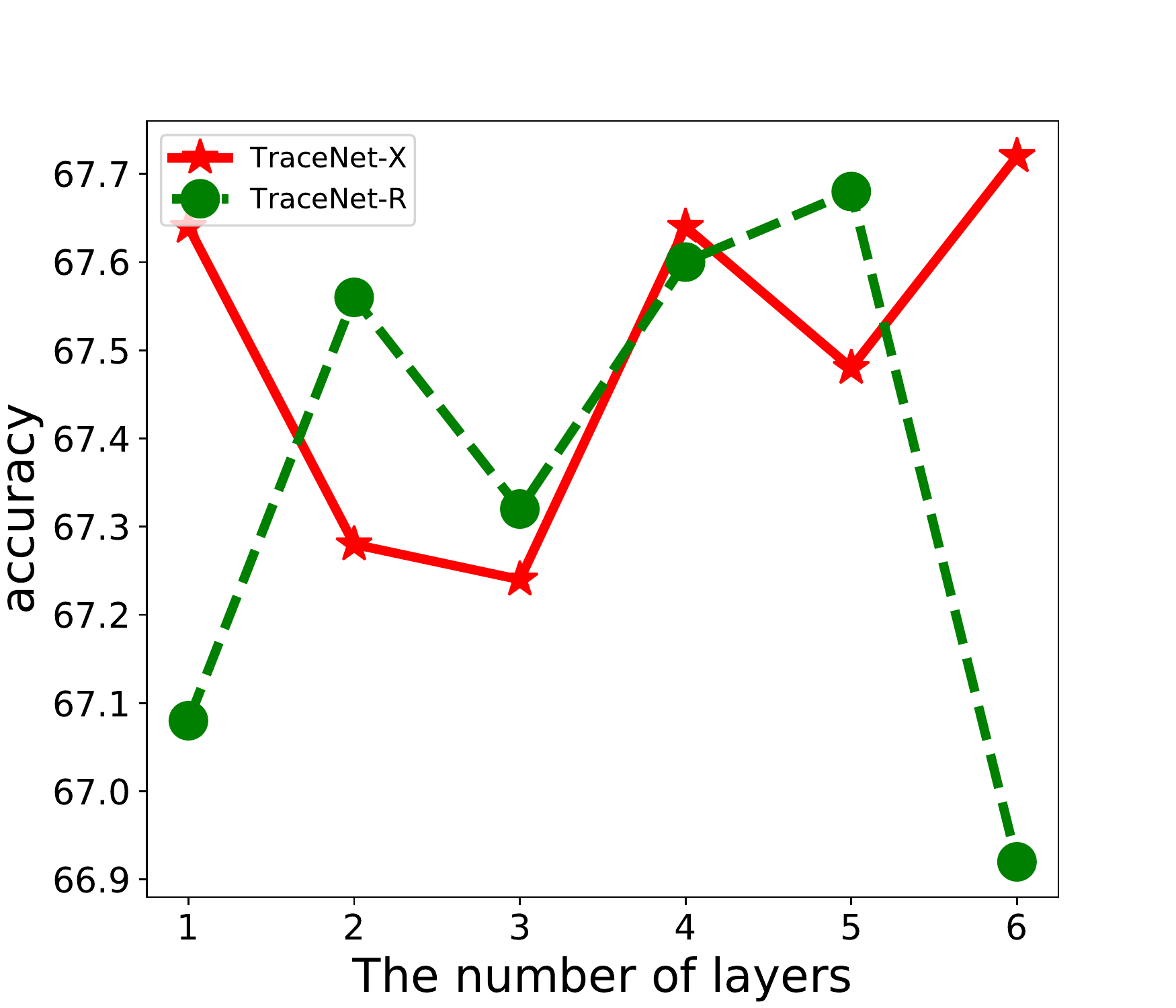}}

\caption{Impacts of the number $L$ of layers (\ie ELCs) on \sstfive.}
\label{exp:sens-ELCs}
\end{figure*}

\subsection{Qualitative Analysis of Item Weights}
We present a qualitative study on item weights estimated in different ELC layer, shown in Fig.~\ref{exp:itemweight}.
The two displayed movie reviews are retrieved from the training set of \sstfive, and their ground-truth sentiment labels are \textit{positive} and \textit{very-positive}, respectively. After training, \modelname could produce accurate labels for both.
In the left case, the key elements identified are \textit{deflated}, \textit{there's much}, and  \textit{recommend}, which make sense for the prediction result. Also note that it remains difficult to find sentiment words at the beginning. However, the multi-layer architecture enables \modelname to eventually refine key elements, \eg \textit{deflated} is identified at the second layer and \textit{recommend} is emphasized finally.
Similarly, \modelname successfully finds the two key words \textit{dark} and \textit{funny} for the right example after learning layer-by-layer.  
To conclude, these item weights generally make the outcomes of \modelname easier to understand.

\subsection{Analysis on Parameter Sensitivity}

\subsubsection{Impacts of masking probability $P_{msk}$}
To evaluate the impacts of $P_{msk}$, we varied $P_{msk}$ from 0 to 1 and computed the classification accuracy of both \modelname-X and  \modelname-R. We omitted \modelname-G since its effectiveness is not comparable to  \modelname-X and  \modelname-R. Each $P_{msk}$ was tested 3 times with different seed, and the averaged value is reported in Fig.~\ref{exp:sens-pmask}.
It turns out that \modelname is quite sensitive to parameter $P_{msk}$, possibly due to the randomness in choosing sentences to mask and choosing masked key items. However, compared with turning off proactive masking (\ie $P_{msk}=0$), our training strategy remains effective within a certain range of $P_{msk}$, \eg $[0.3, 0.5]$ on \sstfive and $[0.05, 0.4]$ on \yelp.

\subsubsection{Impacts of the number $L$ of layers (\ie ELCs)}
To evaluate the impacts of $L$, we varied $L$ from 1 to 6 and computed the \acc of both \modelname-X and  \modelname-R on the two datasets. Note that the discriminator combines all the hidden states to derive the final classification results.  The results are reported in Fig.~\ref{exp:sens-ELCs}.

On the \yelp data, using more layers is generally more effective, while the impacts of $L$ are quite gentle. 
On the other hand, the impacts of $L$ are more complex on the \sstfive data.
When $L \le 3$, the \acc of \modelname increases with the increment of $L$ in general, indicating that \modelname benefits from its multi-layer organization which enables to learn the input structure for multiple times. Further increase $L$ will lead to the decrease of \acc due to over-fitting.
Overall, $L=3$ is a good choice for \modelname, and this conclusion holds for the two variants of \modelname.

\subsubsection{Impacts of hidden state aggregation}

\begin{figure}[htb!]
\centering
\subfigure[\sstfive]{\label{exp:layer-sst}
\includegraphics[width=0.45\columnwidth]{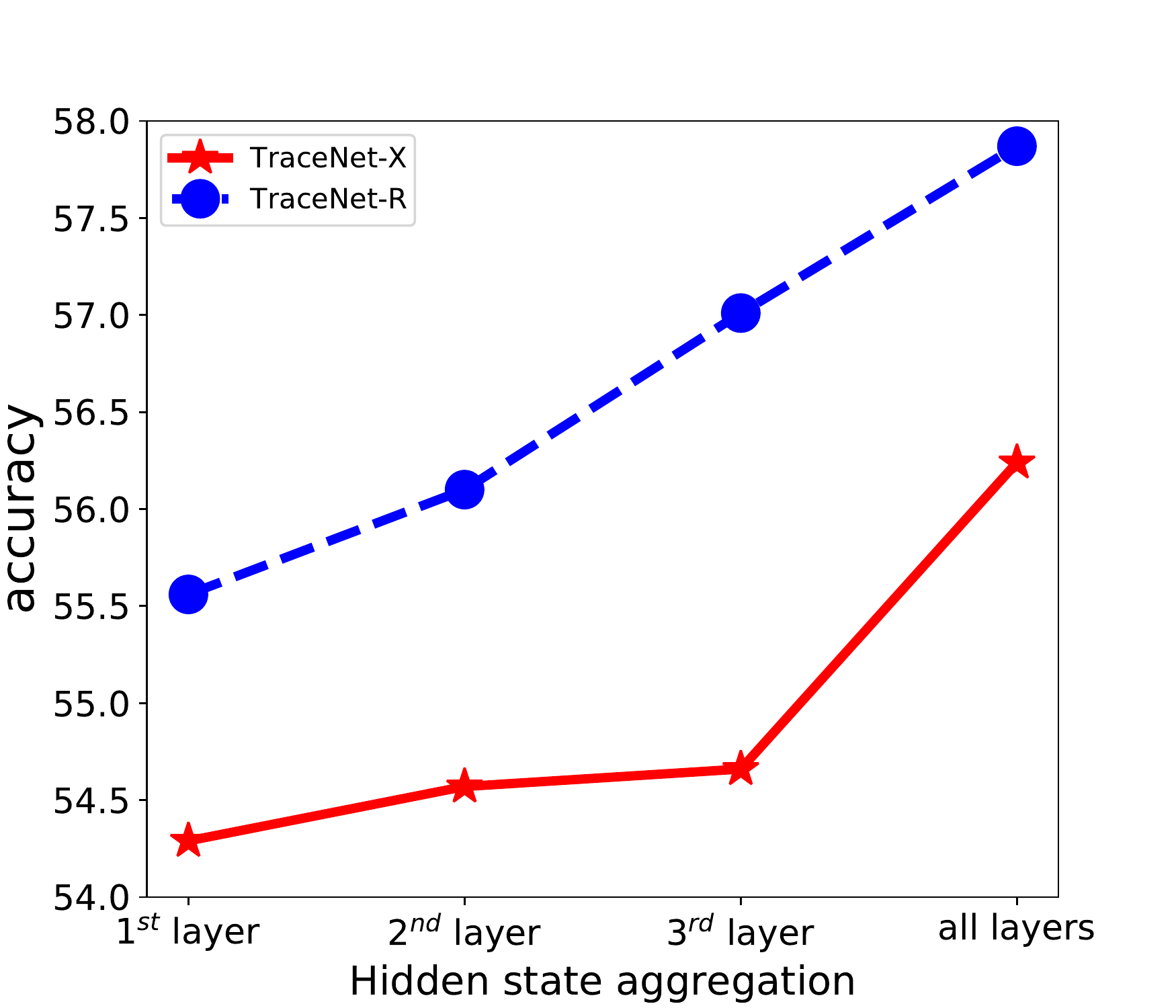}}
\subfigure[\yelp]{\label{exp:layer-yelp}
\includegraphics[width=0.45\columnwidth]{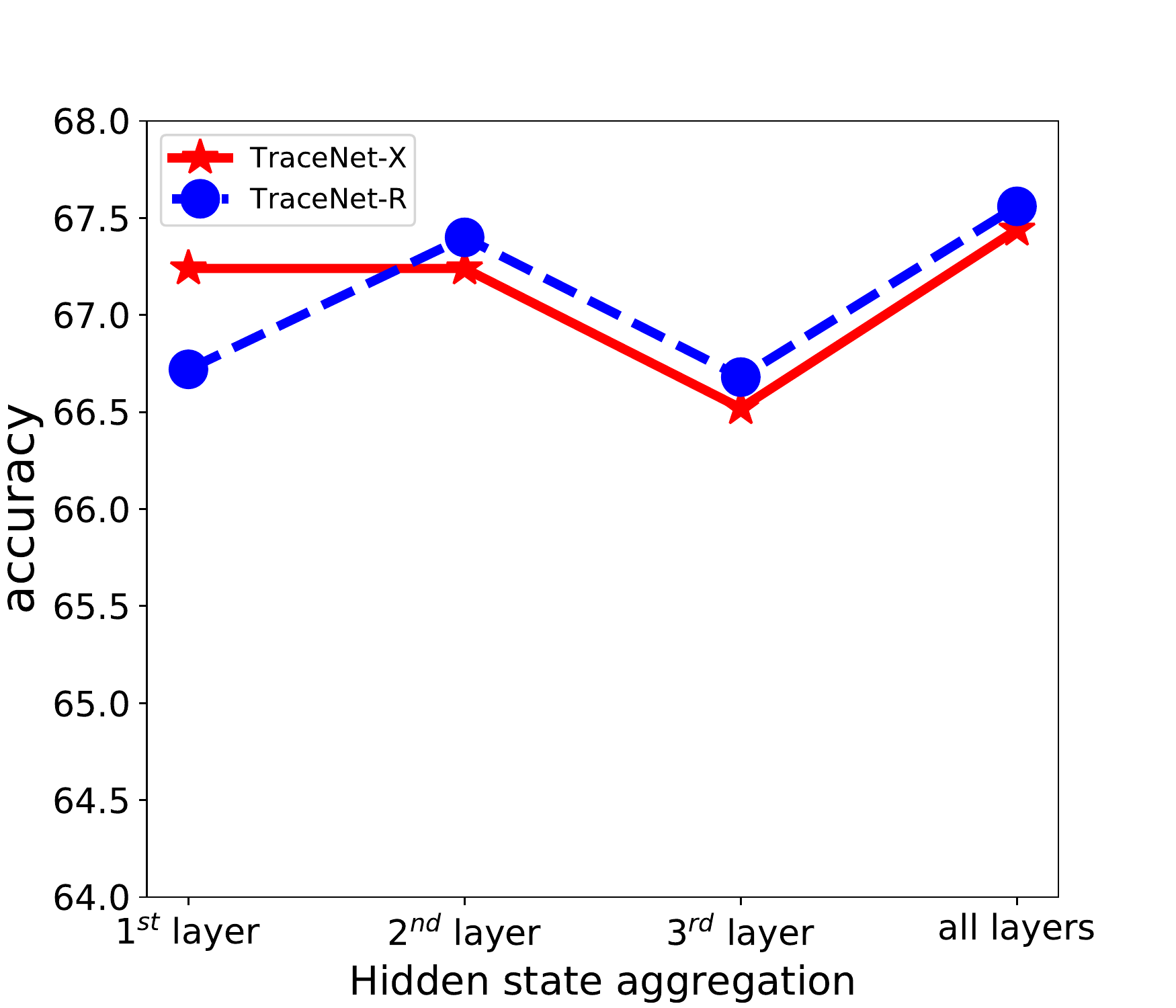}}
\caption{Impacts of hidden state aggregation.}
\label{exp:sens-layer}
\end{figure}

To evaluate the impacts of hidden state aggregation, we computed the \acc of both \modelname-X and  \modelname-R using single hidden states and all the three hidden states on the two datasets. The results are reported in Fig.~\ref{exp:sens-layer}.

For the case of using single hidden states, the best \acc is obtained by the third and second hidden states on the \sstfive and \yelp data, respectively. This is because of their different characteristics of short and long text, \ie the input structure of short sentences is harder to reveal given the limited information than long documents.
Moreover, combining hidden states from all layers is consistently better than using single hidden states alone. We guess that combining hidden states enables the discriminator to directly supervise each layer in terms of revealing the input structure, which enhances the effectiveness.

\section{Conclusion} \label{sec-conc}

In this paper, we proposed \modelname to tackle sentiment analysis task such that the outcome is mainly contributed by a few key elements of the input. The idea behind \modelname, which originates from the two-streams hypothesis, is to learn discriminative representations and reveal input structure simultaneously.
To do this, \modelname stacks several encoders and locators layer-by-layer, with increasing-strength sparsity constraints on locators for tracing key elements. Smoothness regularization is enforced on adjacent encoder-locator layer to ensure the stabilization of learning across layers.
In addition, a proactive masking strategy is further incorporated into \modelname for robustness.
We applied \modelname for sentence- and document-level sentiment analysis. The experiments demonstrated the effectiveness of \modelname. Moreover, considering a total of eight types of attacks, we verified the better robustness of \modelname in general. Finally, our qualitative analysis of item weights showed the advantage of \modelname in terms interpretability.

\bibliography{anthology}

\begin{thebibliography}{26}
\expandafter\ifx\csname natexlab\endcsname\relax\def\natexlab#1{#1}\fi

\bibitem[{Baccianella et~al.(2010)Baccianella, Esuli, and
  Sebastiani}]{baccianella2010swn}
Stefano Baccianella, Andrea Esuli, and Fabrizio Sebastiani. 2010.
\newblock Sentiwordnet 3.0: An enhanced lexical resource for sentiment analysis
  and opinion mining.
\newblock In \emph{Proceedings of the International Conference on Language
  Resources and Evaluation, {LREC}}.

\bibitem[{Bahdanau et~al.(2015)Bahdanau, Cho, and Bengio}]{bahdanau14neural}
Dzmitry Bahdanau, Kyunghyun Cho, and Yoshua Bengio. 2015.
\newblock Neural machine translation by jointly learning to align and
  translate.
\newblock In \emph{3rd International Conference on Learning Representations,
  {ICLR}}.

\bibitem[{Chen and Qian(2019)}]{chen2019transfer}
Zhuang Chen and Tieyun Qian. 2019.
\newblock Transfer capsule network for aspect level sentiment classification.
\newblock In \emph{Proceedings of the 57th Conference of the Association for
  Computational Linguistics, {ACL}}, pages 547--556.

\bibitem[{Cho et~al.(2014)Cho, van Merrienboer, G{\"{u}}l{\c{c}}ehre, Bahdanau,
  Bougares, Schwenk, and Bengio}]{cho2014learning}
Kyunghyun Cho, Bart van Merrienboer, {\c{C}}aglar G{\"{u}}l{\c{c}}ehre, Dzmitry
  Bahdanau, Fethi Bougares, Holger Schwenk, and Yoshua Bengio. 2014.
\newblock Learning phrase representations using {RNN} encoder-decoder for
  statistical machine translation.
\newblock In \emph{Proceedings of the 2014 Conference on Empirical Methods in
  Natural Language Processing, {EMNLP}}, pages 1724--1734.

\bibitem[{Choi et~al.(2018)Choi, Yoo, and Lee}]{choi2018learning}
Jihun Choi, Kang~Min Yoo, and Sang{-}goo Lee. 2018.
\newblock Learning to compose task-specific tree structures.
\newblock In \emph{Proceedings of the Thirty-Second {AAAI} Conference on
  Artificial Intelligence, (AAAI-18)}, pages 5094--5101.

\bibitem[{Devlin et~al.(2019)Devlin, Chang, Lee, and
  Toutanova}]{devlin2019bert}
Jacob Devlin, Ming{-}Wei Chang, Kenton Lee, and Kristina Toutanova. 2019.
\newblock {BERT:} pre-training of deep bidirectional transformers for language
  understanding.
\newblock In \emph{Proceedings of the 2019 Conference of the North American
  Chapter of the Association for Computational Linguistics: Human Language
  Technologies, {NAACL-HLT}}, pages 4171--4186.

\bibitem[{Goodale et~al.(1992)Goodale, Milner et~al.}]{goodale1992separate}
Melvyn~A Goodale, A~David Milner, et~al. 1992.
\newblock Separate visual pathways for perception and action.
\newblock \emph{Trends Neurosci.}, 15(1):20--5.

\bibitem[{Jacobs et~al.(1991)Jacobs, Jordan, and Barto}]{jacobs1991task}
Robert~A Jacobs, Michael~I Jordan, and Andrew~G Barto. 1991.
\newblock Task decomposition through competition in a modular connectionist
  architecture: The what and where vision tasks.
\newblock \emph{Cognitive science}, 15(2):219--250.

\bibitem[{Johnson and Zhang(2015)}]{johnson2015effective}
Rie Johnson and Tong Zhang. 2015.
\newblock Effective use of word order for text categorization with
  convolutional neural networks.
\newblock In \emph{{NAACL} {HLT} 2015, The 2015 Conference of the North
  American Chapter of the Association for Computational Linguistics: Human
  Language Technologies}, pages 103--112.

\bibitem[{Kim(2014)}]{kim2014convolutional}
Yoon Kim. 2014.
\newblock Convolutional neural networks for sentence classification.
\newblock In \emph{Proceedings of the 2014 Conference on Empirical Methods in
  Natural Language Processing, {EMNLP}}, pages 1746--1751.

\bibitem[{Liu et~al.(2019)Liu, Ott, Goyal, Du, Joshi, Chen, Levy, Lewis,
  Zettlemoyer, and Stoyanov}]{liu2019roberta}
Yinhan Liu, Myle Ott, Naman Goyal, Jingfei Du, Mandar Joshi, Danqi Chen, Omer
  Levy, Mike Lewis, Luke Zettlemoyer, and Veselin Stoyanov. 2019.
\newblock Roberta: {A} robustly optimized {BERT} pretraining approach.
\newblock \emph{CoRR}, abs/1907.11692.

\bibitem[{Madotto et~al.(2018)Madotto, Wu, and Fung}]{Fung2018mem2seq}
Andrea Madotto, Chien{-}Sheng Wu, and Pascale Fung. 2018.
\newblock Mem2seq: Effectively incorporating knowledge bases into end-to-end
  task-oriented dialog systems.
\newblock In \emph{Proceedings of the 56th Annual Meeting of the Association
  for Computational Linguistics, {ACL}}, pages 1468--1478.

\bibitem[{Martins and Astudillo(2016)}]{martins2016sparsemax}
Andr{\'{e}} F.~T. Martins and Ram{\'{o}}n~Fern{\'{a}}ndez Astudillo. 2016.
\newblock From softmax to sparsemax: {A} sparse model of attention and
  multi-label classification.
\newblock In \emph{Proceedings of the 33nd International Conference on Machine
  Learning, {ICML}}, volume~48, pages 1614--1623.

\bibitem[{Mikolov et~al.(2013)Mikolov, Sutskever, Chen, Corrado, and
  Dean}]{mikolov2013word2vec}
Tomas Mikolov, Ilya Sutskever, Kai Chen, Gregory~S. Corrado, and Jeffrey Dean.
  2013.
\newblock Distributed representations of words and phrases and their
  compositionality.
\newblock In \emph{Advances in Neural Information Processing Systems 26}, pages
  3111--3119.

\bibitem[{Pennington et~al.(2014)Pennington, Socher, and
  Manning}]{pennington2014glove}
Jeffrey Pennington, Richard Socher, and Christopher~D. Manning. 2014.
\newblock Glove: Global vectors for word representation.
\newblock In \emph{Proceedings of the 2014 Conference on Empirical Methods in
  Natural Language Processing, {EMNLP}}, pages 1532--1543.

\bibitem[{Simonyan and Zisserman(2014)}]{simonyan2014two}
Karen Simonyan and Andrew Zisserman. 2014.
\newblock Two-stream convolutional networks for action recognition in videos.
\newblock \emph{arXiv preprint arXiv:1406.2199}.

\bibitem[{Socher et~al.(2013)Socher, Perelygin, Wu, Chuang, Manning, Ng, and
  Potts}]{socher2013sst5}
Richard Socher, Alex Perelygin, Jean Wu, Jason Chuang, Christopher~D. Manning,
  Andrew~Y. Ng, and Christopher Potts. 2013.
\newblock Recursive deep models for semantic compositionality over a sentiment
  treebank.
\newblock In \emph{Proceedings of the 2013 Conference on Empirical Methods in
  Natural Language Processing, {EMNLP}}, pages 1631--1642.

\bibitem[{Sukhbaatar et~al.(2015)Sukhbaatar, Weston, Fergus
  et~al.}]{sukhbaatar2015end}
Sainbayar Sukhbaatar, Jason Weston, Rob Fergus, et~al. 2015.
\newblock End-to-end memory networks.
\newblock In \emph{Advances in neural information processing systems}, pages
  2440--2448.

\bibitem[{Tang et~al.(2019)Tang, Lu, Su, Ge, Song, Sun, and
  Luo}]{tang2019progressive}
Jialong Tang, Ziyao Lu, Jinsong Su, Yubin Ge, Linfeng Song, Le~Sun, and Jiebo
  Luo. 2019.
\newblock Progressive self-supervised attention learning for aspect-level
  sentiment analysis.
\newblock In \emph{Proceedings of the 57th Conference of the Association for
  Computational Linguistics, {ACL}}, pages 557--566.

\bibitem[{Vaswani et~al.(2017)Vaswani, Shazeer, Parmar, Uszkoreit, Jones,
  Gomez, Kaiser, and Polosukhin}]{vaswani2017attention}
Ashish Vaswani, Noam Shazeer, Niki Parmar, Jakob Uszkoreit, Llion Jones,
  Aidan~N Gomez, {\L}ukasz Kaiser, and Illia Polosukhin. 2017.
\newblock Attention is all you need.
\newblock In \emph{Advances in neural information processing systems}, pages
  5998--6008.

\bibitem[{Velickovic et~al.(2018)Velickovic, Cucurull, Casanova, Romero,
  Li{\`{o}}, and Bengio}]{velickovic18gat}
Petar Velickovic, Guillem Cucurull, Arantxa Casanova, Adriana Romero, Pietro
  Li{\`{o}}, and Yoshua Bengio. 2018.
\newblock Graph attention networks.
\newblock In \emph{6th International Conference on Learning Representations,
  {ICLR}}.

\bibitem[{Wang and Liu(2018)}]{8630333}
Jiangliu Wang and Yunhui Liu. 2018.
\newblock \href {https://doi.org/10.1109/WCICA.2018.8630333} {Kinematics
  features for 3d action recognition using two-stream cnn}.
\newblock In \emph{2018 13th World Congress on Intelligent Control and
  Automation (WCICA)}, pages 1731--1736.

\bibitem[{Yang et~al.(2019)Yang, Dai, Yang, Carbonell, Salakhutdinov, and
  Le}]{Yang2019xlnet}
Zhilin Yang, Zihang Dai, Yiming Yang, Jaime~G. Carbonell, Ruslan Salakhutdinov,
  and Quoc~V. Le. 2019.
\newblock Xlnet: Generalized autoregressive pretraining for language
  understanding.
\newblock In \emph{Advances in Neural Information Processing Systems 32}, pages
  5754--5764.

\bibitem[{Zhang et~al.(2018)Zhang, Wang, and Liu}]{zhang2018deep}
Lei Zhang, Shuai Wang, and Bing Liu. 2018.
\newblock Deep learning for sentiment analysis: A survey.
\newblock \emph{Wiley Interdisciplinary Reviews: Data Mining and Knowledge
  Discovery}, 8(4):e1253.

\bibitem[{Zhang et~al.(2021)Zhang, Liu, Wang, Zeng, and Mei}]{zhang2021robust}
Ning Zhang, Jingen Liu, Ke~Wang, Dan Zeng, and Tao Mei. 2021.
\newblock Robust visual object tracking with two-stream residual convolutional
  networks.
\newblock In \emph{2020 25th International Conference on Pattern Recognition
  (ICPR)}, pages 4123--4130. IEEE.

\bibitem[{Zhang and Goldwasser(2019)}]{zhang2019sentiment}
Xiao Zhang and Dan Goldwasser. 2019.
\newblock Sentiment tagging with partial labels using modular architectures.
\newblock In \emph{Proceedings of the 57th Conference of the Association for
  Computational Linguistics, {ACL}}, pages 579--590.

\end{thebibliography}
\bibliographystyle{acl_natbib}

\appendix

\section{Example Appendix}

\begin{table*}[h]
\centering
\label{table:sst5}
\begin{small}
\begin{tabular}{m{1.7cm} | m{5.5cm}  | m{5.5cm} } \hline
\bf{Algorithm} & \sstfive & \yelp \\ \hline
 \cnnrandom \newline \cnnstatic \newline  \cnnnons \newline  \cnnmulch & kernel size: \{2,3,4,5\} \newline filter number (per kernel size): 300  \newline  $L_2$ weight: 0.01 \newline batch size: 50 \newline learning rate: 0.001 \newline sequence length: 49 & kernel size: \{2,3,4,5\} \newline filter number (per kernel size): 300 \newline  $L_2$ weight: 0.01 \newline batch size: 50 \newline learning rate: 0.001 \newline sequence length: 256 \\\hline
 
 \lstm \newline \bilstm & hidden state size: 100 \newline $L_2$ weight: 0.01 \newline batch size: 50 \newline learning rate: 0.001 \newline sequence length: 49 \newline dropout: 0.5 & hidden state size: 100 and 50, respectively \newline $L_2$ weight: 0.01 \newline  batch size: 50 \newline learning rate: 0.001 \newline sequence length: 256 \newline dropout: 0.5 \\ \hline
 
 \treelstm & hidden state size: 300 \newline batch size: 64 \newline learning rate: 1.0, halved every two epochs \newline sequence length: 49 \newline dropout: 0.5 & hidden state size: 300 \newline batch size: 16 \newline learning rate: 1.0, halved every two epochs \newline sequence length: 256 \newline dropout: 0.5. \\ \hline
 
 \bert \newline \xlnet \newline \roberta & hidden state size: 768 \newline model type: base-cased, base and base, resp. \newline weight decay: 0.1, 0.1, and 0.0, resp. \newline Adam epsilon: 1e-8, 1e-8, and 1e-6, resp. \newline batch size: 32, 16, and 16, resp. \newline learning rate: 5e-5, 2e-5, and 2e-5, resp. \newline sequence length: 128, 64, and 128, resp. \newline dropout 0.1 & hidden state size: 768 \newline model type: base-cased, base and base, resp. \newline weight decay: 0.1, 0.1 and 0.0, resp. \newline Adam epsilon: 1e-8, 1e-8, and 1e-6, resp. \newline batch size: 64 \newline learning rate: 5e-5, 2e-5, and 2e-5, resp. \newline sequence length: 256 \newline dropout: 0.1  \\ \hline
 
\modelname-G \newline \modelname-X \newline \modelname-R   & hidden state size: 50, 128, and 512, resp.  \newline weight decay: 0.2, 0.1, and 0.0, respectively \newline Adam epsilon: 1e-8, 1e-8, and 1e-6, resp. \newline batch size: 64, 16, and 16, respectively \newline learning rate: 1e-3, 2e-5,  and 2e-5, resp. \newline sequence length: 49, 64, and 128, respectively \newline dropout: 0.2, 0.3, and 0.1, respectively \newline $P_{msk}$: 0.05, 0.2 and 0.3, respectively  \newline number $L$ of layers: 3 & hidden state size: 500, 512, and 768, resp.  \newline weight decay: 0.2, 0.1, and 0.1, respectively  \newline Adam epsilon: 1e-8 \newline batch size: 64 \newline learning rate: 1e-3, 2e-5, and 2e-5, resp. \newline sequence length: 256 \newline dropout: 0.2, 0.1, and 0.1, respectively. $P_{msk}$: 0.05, 0.05 and 1.0, respectively \newline number $L$ of layers: 3 \\ \hline
\end{tabular}
\caption{Hyper-parameters setting. (49 refers to the maximum sentence length of \sstfive.)}
\end{small}
\end{table*}

\section*{Appendix A: Experimental Details} \label{sec-app-exp-details}

We first present more experimental details for reproduce purpose.

Public \sstfive\footnote{\url{https://nlp.stanford.edu/sentiment/}} and \yelp\footnote{\url{http://goo.gl/JyCnZq}} datasets are choosed to evaluate our \modelname architecture.
We adopted a third-party implementation\footnote{{\url{https://github.com/andyweizhao/capsule_text_classification}}} for  \cnnrandom, \cnnstatic, \cnnnons, \cnnmulch, \lstm, and \bilstm. 
The source code of \treelstm had been released\footnote{\url{https://github.com/jihunchoi/unsupervised-treelstm}} by its authors. We implemented \bert, \xlnet, \roberta  and \modelname based on Hugging Face library\footnote{\url{https://github.com/huggingface/transformers}}. All hyper-parameters of these approaches are summarized in Table~1.
Finally, when initializing word embedding with pretrained vectors, glove.840B.300d\footnote{\url{https://nlp.stanford.edu/projects/glove/}} is adopted. Words not in the pretrained vectors vocabulary are initialized randomly. 
We have attached the code and data in the supplementary material.

All our tests were performed on Tesla V100 GPUs with 32GB memory. 
Model selection was performed according to the performance on the validation set such that the CNN- and LSTM-based baselines were trained for a maximum of 20 epochs and the rest approaches for a maximum of 10 epochs.

\end{document}